\documentclass[conference]{IEEEtran}
\IEEEoverridecommandlockouts
\usepackage{cite}
\usepackage{amsmath,amssymb,amsfonts}
\usepackage{graphicx}
\usepackage{booktabs}
\usepackage{multirow}
\usepackage{xcolor}
\usepackage{tikz}
\usetikzlibrary{arrows.meta,positioning,fit,backgrounds}
\usepackage[hidelinks]{hyperref}
\hypersetup{pdftitle={AMBIT: Anticipatory Multimodal Body Recruitment for Bimanual Tracking on a Humanoid}}
\graphicspath{{figures/}}
\newcommand{\todo}[1]{\textcolor{red}{[TODO: #1]}}
\newcommand{\eps}{\varepsilon}
\newcommand{\Zvalid}{\mathcal{Z}_{\mathrm{valid}}}
\newcommand{\tsat}{t_{\mathrm{sat}}}
\newcommand{\trec}{t_{\mathrm{rec}}}

\begin{document}
\bstctlcite{IEEEexample:BSTcontrol}

\title{AMBIT: Anticipatory Multimodal Body Recruitment for Bimanual Tracking on a Humanoid}

\author{\IEEEauthorblockN{Hanlong Li, Sihan Tan, Takeshi Ashizawa, Benjamin Yen, and Kazuhiro Nakadai}
\IEEEauthorblockA{Institute of Science Tokyo, Tokyo, Japan}}

\maketitle

\begin{figure*}[t]
\centering
\IfFileExists{figures/fig1_overview.pdf}{\includegraphics[width=0.82\textwidth]{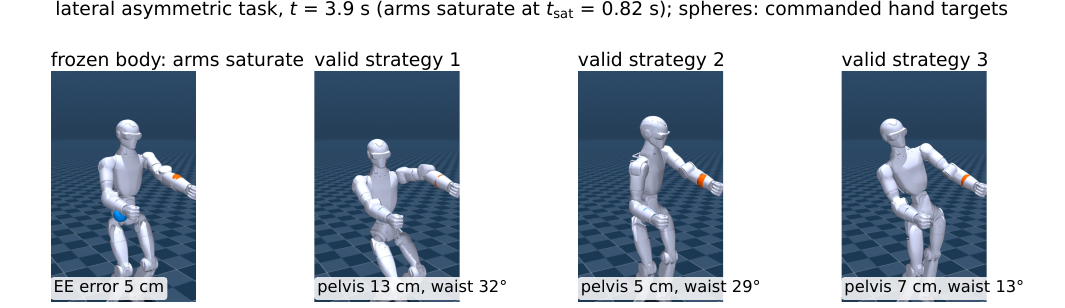}}{\todo{overview figure missing; run the figure-1 script}}
\caption{One bimanual task, four bodies. With pelvis and waist frozen the 5-DoF arms saturate and the hands miss
their targets (left); the same task admits several dynamically valid \emph{recruitment strategies} that lean,
lower or twist the body (right; MuJoCo, Unitree R1). We learn their conditional
distribution, select among samples with a non-learned verifier, and execute the chosen one as the reference of a
standard whole-body tracker.}
\label{fig:overview}
\end{figure*}

\begin{abstract}
A humanoid with 5-DoF arms cannot track generic bimanual end-effector trajectories with its arms alone; pelvis and
waist motion must be recruited, but which motion to use, and when, is not uniquely determined. We show on a Unitree
R1 in fixed double support that the set of dynamically valid \emph{recruitment strategies} (pelvis pose and waist
trajectories) for a task is a diverse continuous manifold, and that a deterministic regressor trained on it
mode-averages into strategies valid only 35\% of the time, against 52\% for a conditional variational autoencoder
(CVAE) and 82\% for the best of 16 CVAE samples. We therefore introduce AMBIT, a generative recruitment planner: the
CVAE proposes strategies conditioned on a preview of the commanded trajectory; a non-learned selector filters, ranks,
and verifies them; and a receding-horizon loop commits to one with hysteresis. The committed strategy is executed as
the reference for the same whole-body differential-IK QP a reactive tracker runs, so the tracker keeps authority over
residual end-effector error. Across 160 held-out episodes that admit a valid strategy, in full MuJoCo dynamics under
a whole-body torque controller, AMBIT reaches 85\% success at a 3\,cm/15$^\circ$ tolerance against 74\% for the
reactive tracker, with non-overlapping confidence intervals. AMBIT also recruits the body before the arms saturate
in 48\% of episodes against 35\%. Because the learned distribution preserves diversity, constraints unknown at
training time are enforced at test time by selection alone. Under five zero-shot shifts, AMBIT outperforms the
warm-started tracker on every shift. It also matches a test-time re-optimisation baseline 17$\times$ more expensive.
Additionally, the entire protocol on a Unitree G1, with all hyperparameters unchanged, reproduces the structure of
the valid set and widens the gap over the tracker to 0.85 against 0.53. Finally, five selected strategies execute on
the externally supported physical R1, staying distinct in pelvis excursion and tracking the planned end-effector
motion to a median of 11\,mm by encoder forward kinematics, which establishes kinematic realisability, not balance.
\end{abstract}

\section{Introduction}
Humanoid arms have too few joints. On many current platforms each arm has five, so the arm alone cannot realise an
arbitrary end-effector (EE) pose in $SE(3)$, and bimanual tasks a human would do with the arms require the robot
to move its trunk, pelvis and legs. Whole-body controllers handle this reactively: a task-priority quadratic program (QP) tracks the
EE targets and a low-priority posture cost decides how the body gives way~\cite{li2025amo}. This hides two
choices. \emph{Which} body motion to recruit is not unique: leaning, lowering the pelvis, twisting the waist
and shifting laterally can all realise the same hand poses (Fig.~\ref{fig:overview}), and their validity depends
on balance, joint limits and actuator torque along the whole trajectory. And \emph{when}: a reactive tracker moves
the body only once the arms have saturated, which is late, and the body is slow.

This paper treats body recruitment as a \emph{temporally extended, multimodal} decision. A recruitment strategy
$z$ is the pelvis pose and waist trajectory over the task horizon, and the valid set $\Zvalid$ holds the
strategies whose full-dynamics execution completes the task within tolerance without falling, slipping, colliding
or exceeding limits. Four questions, each tied to an experiment (Sec.~\ref{sec:exp}): (H1) is $\Zvalid$
meaningfully multimodal at realistic tolerances; (H2) does deterministic regression onto it mode-average into
degraded strategies; (H3) does previewing the commanded trajectory let a learned planner recruit the body earlier
than a reactive tracker; (H4) does a preserved distribution let constraints unknown at training time be honoured
at test time by selection alone.

\textbf{Contributions.}
(i) A formulation and an oracle dataset of dynamically validated recruitment strategies for a Unitree R1 (3000
episodes, 11 valid strategies each) answering H1 and H2: the valid set is a diverse continuous manifold of intrinsic
dimension about four, and regression onto it yields strategies valid 35\% of the time against 52\% for a CVAE.
(ii) AMBIT, a generative recruitment planner whose sampled strategies are executed as \emph{references of a standard
whole-body QP tracker}, so the learned component decides how and when the body is recruited while the tracker
keeps reactive authority over residual hand error. Against the same tracker with a fixed home-posture reference it
is significantly better at both tolerances (85\% vs.\ 74\% and 74\% vs.\ 51\%) and recruits earlier (48\% vs.\
35\% of episodes).
(iii) Zero-shot constraint transfer (H4): five shifts, including a wrist payload, enforced at test time by
filtering and verifying samples; the planner beats the warm-started tracker on every shift and matches a 25\,s
random-restart re-optimisation baseline (B4) at 1.5\,s.
(iv) The same protocol on a Unitree G1, hyperparameters unchanged: the structural findings replicate (9.0 distinct
strategies per solvable episode, intrinsic dimension 4.3, regressor 0.28 against CVAE 0.47) and the margins grow
(0.85 against 0.53 closed loop; up to 0.90 against 0.53 under shift, beating B4 on all five).
(v) Five selected strategies executed on the physical R1 with the robot externally supported: commanded to 4.0, 5.4
and 3.0\,cm of peak pelvis excursion, they measure 3.5, 4.7 and 2.7\,cm and reproduce the planned end-effector
motion to a median of 11\,mm by forward kinematics of the joint encoders (no external end-effector measurement).
The runs establish kinematic realisability, not dynamic balance (Sec.~\ref{sec:hardware}).

We also report what did not work as hoped: anticipation margins are tens of milliseconds, because arm saturation
comes early (median 0.7\,s) and valid recruitment is capped at 0.15\,m/s; preview conditioning is nonetheless
necessary (56\% success without it against 82\%).

\section{Related Work}
\textbf{Distributions over inverse-kinematics (IK) solutions.} IKFlow~\cite{ames2022ikflow} and GGIK~\cite{limoyo2023ggik} learn
generative models over joint configurations for a fixed-base manipulator and evaluate them by precision and by
the maximum mean discrepancy (MMD)~\cite{gretton2012mmd}; IKDiffuser~\cite{ikdiffuser2025} and whole-body IK
graph diffusion~\cite{wbikdiffusion2026} extend the idea to redundant and floating-base kinematic chains. These
models are \emph{per pose}: one configuration per target, with temporal consistency, dynamics and balance left to
a downstream controller. Our strategies are trajectories over the whole task horizon, validated in dynamics; the per-pose baseline B3
(Sec.~\ref{sec:exp}) chatters between modes (46 switches per episode) and cannot anticipate.

\textbf{Whole-body control and motion generation.} Whole-body recruitment for end-effector tasks under balance
and physical constraints is classical, not a contribution of this paper: resolved momentum
control generates whole-body motion from desired linear and angular momentum~\cite{kajita2003rmc}, the dynamics
filter turns physically inconsistent input motion into dynamically consistent whole-body motion
online~\cite{yamane2003dynamicsfilter}, whole-body tracking controllers reproduce captured human motion with all
joints while keeping balance~\cite{nakaoka2003dances,yamane2009tracking}, and task-priority control and whole-body
motion planning integrate manipulation objectives with balance, contact, joint-limit and collision
constraints~\cite{sentis2005synthesis,kanoun2011prioritized,dalibard2013wbmp,yoshida2017wbmp}. Learned controllers
such as ExBody~\cite{cheng2024exbody}, HumanPlus~\cite{fu2024humanplus}, OmniH2O~\cite{he2024omnih2o} and
AMO~\cite{li2025amo} track a whole-body or upper-body reference from a human or a motion prior. All of these produce
or execute \emph{one} feasible motion; ours is complementary: for a given bimanual trajectory we learn the conditional distribution of the \emph{many} dynamically valid, temporally extended recruitment
strategies, preview the future task, and select among samples under constraints introduced only at test time; the
chosen strategy becomes the reference of a standard whole-body QP tracker, which keeps reactive authority over
residual hand error. Our reactive baseline B1 is that same QP with a fixed home-posture reference, so the comparison
isolates the reference.

\textbf{Mode averaging.} Diffusion Policy~\cite{chi2023diffusionpolicy} popularised the argument that
regressing a multimodal action distribution produces invalid averages. We test the argument for recruitment
strategies with a matched-data regressor (B2) and quantify the gap in dynamic validity rather than in imitation
loss.

\section{Problem Formulation}\label{sec:problem}
\textbf{Platform and task.} The Unitree R1 has 24 body joints: six per leg, two at the waist (roll and yaw, no
pitch) and five per arm (shoulder pitch/roll/yaw, elbow, wrist roll). We consider bimanual tracking of a commanded
EE trajectory $\tau_{EE}=\{T^*_L(t),T^*_R(t)\}\in SE(3)\times SE(3)$ over $t\in[0,T]$, $T\in[3,6]$\,s, in fixed
double support without stepping or object contact. The task is solved within tolerance
$\eps=(\eps_p,\eps_R)$ if both hands stay within $\eps_p$ in position and $\eps_R$ in orientation of their
targets at every control step. We report a strict level (2\,cm, 10$^\circ$) and a mild level (3\,cm, 15$^\circ$);
a loose level (5\,cm, 20$^\circ$) appears only in Table~\ref{tab:e2}.

\textbf{Why the body must be recruited.} A single 5-DoF arm has a Jacobian of rank 5 at every one of 2000 sampled
configurations, so one arm cannot track a generic $SE(3)$ target. The ten arm joints and the two waist joints
form a $12\times12$ system for the 12-D bimanual task that is full rank in all samples but frequently near
singular (median smallest singular value $5\times10^{-3}$). Adding the six pelvis degrees of freedom, realised
by the legs with the feet fixed, yields an exact six-dimensional null space: the space of recruitment strategies. With pelvis and waist frozen at the initial configuration and the arms tracking by IK,
99\% of our task episodes exceed the strict tolerance at some time $\tsat$ (defined below); recruitment is
necessary, not optional.

\textbf{Recruitment strategies and the valid set.} A strategy $z_{0:T}$ is the sequence of pelvis pose (position and orientation) and the two waist angles at a 50\,ms timestep. Given $z$, the legs are determined
by per-leg IK from the pelvis pose with the feet fixed, and the arms are resolved against $\tau_{EE}$ by a
differential-IK QP (Sec.~\ref{sec:method}). The valid set $\Zvalid(\tau_{EE};\eps)$ contains the strategies whose
\emph{full-dynamics} execution (Sec.~\ref{sec:exec}) satisfies the EE tolerance at every step, keeps the centre of
mass (CoM) inside the support polygon, neither falls nor slips (foot displacement below 2\,cm), avoids self-collision,
respects joint position and velocity limits and saturates actuator torque in at most 5\% of steps.

\textbf{Definitions used throughout.} The \emph{frozen-body counterfactual} rolls out the arms alone with the body
frozen; $\tsat$ is the first time its EE error exceeds the strict tolerance. The \emph{recruitment onset}
$\trec$ of an executed motion is the first time the pelvis deviates from its initial pose by more than 2\,cm or
$3^\circ$, or the waist by more than $3^\circ$. The \emph{anticipation margin} is $\Delta t=\tsat-\trec$; positive
means the body moved before the arms saturated.


\section{Method}\label{sec:method}
\begin{figure}[t]
\centering
\resizebox{\columnwidth}{!}{%
\begin{tikzpicture}[
  font=\scriptsize, >=Latex,
  box/.style={draw, rounded corners=1pt, align=center, minimum height=5.5mm, inner sep=2pt, fill=white},
  learned/.style={box, fill=blue!8},
  analytic/.style={box, fill=orange!10},
  ctrl/.style={box, fill=green!8},
  node distance=2.2mm and 3mm]
\node[box] (in) {commanded $\tau_{EE}$ (preview)\\+ body state $s_t$};
\node[learned, right=of in] (cvae) {CVAE sampler\\$K$ strategies $z^{(k)}$};
\node[analytic, right=of cvae] (filt) {z-space filters\\(shift constraints)};
\node[analytic, right=of filt] (cost) {analytic cost\\$C(z)$};
\node[analytic, below=of cost] (ver) {kinematic verification\\(arms-only QP, 1\,s)};
\node[analytic, left=of ver] (commit) {commitment\\(hysteresis, blend)};
\node[ctrl, left=of commit] (qp) {whole-body QP tracker\\ref.\ $=z$ (soft) / home (fallback)};
\node[ctrl, left=of qp] (tq) {torque QP\\+ robot (MuJoCo)};
\draw[->] (in) -- (cvae); \draw[->] (cvae) -- (filt); \draw[->] (filt) -- (cost); \draw[->] (cost) -- (ver);
\draw[->] (ver) -- (commit); \draw[->] (commit) -- (qp); \draw[->] (qp) -- (tq);
\draw[->, dashed] ([xshift=1.2mm]tq.north west) |- (in.west) node[pos=0.75, below, font=\tiny, align=center, inner sep=1pt]{replan\\every 0.5\,s};
\end{tikzpicture}}
\caption{Pipeline. Blue: learned; orange: non-learned selection; green: control. B1 is the same whole-body QP
tracker with the home posture as reference.}
\label{fig:pipeline}
\end{figure}
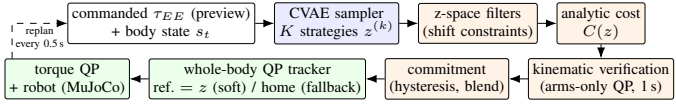

Fig.~\ref{fig:pipeline} summarises AMBIT: an offline oracle produces diverse validated strategies
(Sec.~\ref{sec:oracle}), a CVAE learns their conditional distribution (Sec.~\ref{sec:model}), a non-learned
selector filters, ranks and verifies $K$ samples at run time (Sec.~\ref{sec:selector}), and a receding-horizon
loop commits to one and executes it as the reference of a whole-body QP tracker (Sec.~\ref{sec:runtime}) under a
whole-body torque controller (Sec.~\ref{sec:exec}).

\subsection{Oracle and dataset}\label{sec:oracle}
For each episode we generate 41 candidate strategies with the whole-body QP: one \emph{reactive} strategy (home
posture reference, so the body moves only when the arms saturate), eight \emph{anticipatory} time-warps of it
that start earlier, eight null-space samples tracking randomised smooth pelvis and waist references, and sixteen
Sobol samples over the QP's per-group posture and motion weights. Every candidate is executed in full dynamics
and labelled valid per tolerance. Valid candidates are deduplicated by farthest-point selection in a descriptor
space (pelvis translation and rotation, waist use, CoM displacement, recruitment onset; scales 5\,cm, 5$^\circ$,
5$^\circ$, 3\,cm, 0.25\,s); the number of representatives is the episode's mode count.

The task suite has five families: forward reach-and-hold, low targets below pelvis height, lateral and asymmetric
bimanual targets, traversals from an easy into a hard region, and constant-relative-pose transport arcs. The main shard has 3000 episodes; a second shard of 2000 episodes with narrower parameter
ranges is used only to test generalisation. One physics finding of the oracle shapes everything downstream
(Fig.~\ref{fig:e0}): dynamic validity collapses when the planned pelvis speed exceeds about 0.2\,m/s, and oracle
strategies that recruit the body more than 0.3\,s before the arms saturate are valid 60\% of the time against
46\% for those that recruit after. Early recruitment is slow recruitment.

\subsection{Generative model}\label{sec:model}
We represent a strategy as the deviation of eight pelvis and waist coordinates (position, rotation vector, waist
angles) from the home configuration, compressed with a 16-coefficient discrete cosine transform (DCT) per
coordinate (128 dimensions). The condition is the
DCT of the commanded EE trajectory expressed relative to the initial hand poses (192 dimensions), the remaining
duration, and an 8-D body-state deviation from home. Because the valid set is a continuous manifold rather than a
set of discrete modes (Sec.~\ref{sec:exp}), the model is a CVAE~\cite{sohn2015cvae} with a 16-D latent, trained
with a min-over-5-samples reconstruction loss to resist posterior collapse and $\beta=10^{-2}$. A conditional
affine-coupling flow~\cite{dinh2017realnvp} serves as a second sampler; the B2 baseline is an MLP regressor with
the same inputs and an MSE loss. For the
receding-horizon runtime the condition is computed over the remaining window at every replan, so the model
previews the future commanded trajectory. The \emph{no-preview} ablation uses the same architecture, data and
runtime but replaces future targets by the current target.

\subsection{Selection}\label{sec:selector}
Selection is non-learned. The $K$ samples are first checked in strategy space against hard geometric
constraints (pelvis keep-out boxes, torso-obstacle clearance, waist-range limits, forbidden lean half-spaces); this
is where test-time constraint shifts enter (Sec.~\ref{sec:e3}). Survivors are ranked by an analytic cost
$C(z)=w_eC_{EE}+w_mC_{\mathrm{motion}}+w_bC_{\mathrm{balance}}+w_cC_{\mathrm{constraint}}$ with the time-mean
EE error, a mean-square pelvis and waist velocity, the shortfall of the CoM margin below 4\,cm and a soft
constraint penalty. Strategy-space features predict dynamic validity poorly, so the leading candidates are
\emph{verified} in cost order by an arms-only QP pass over the next second of the plan
(pass if both hands stay within 2\,cm for the soft executor of Sec.~\ref{sec:runtime}); the first to pass is
chosen, at most six are tried. With a wrist payload the pass adds a quasi-static CoM check.

\subsection{Runtime: commitment and soft execution}\label{sec:runtime}
The planner replans every 0.5\,s with $K=16$ samples. The incumbent strategy's remaining segment is always a
candidate; the planner switches only if the incumbent fails verification or a challenger improves the cost by
more than 15\%, and blends references over 0.3\,s on a switch. If no candidate passes verification, the tracker
falls back to the reactive reference (home posture) until the next replan. \emph{Planning latency is measured but
not simulated.} One plan takes 1.5\,s of wall time in our Python implementation, three times the replan interval;
sampling and strategy-space checks take milliseconds, the verification QP over the next second of the plan
dominates. The rollout clock pauses while the planner runs, so the closed-loop numbers below are those of a
zero-latency planner. Replanning only every 1.5\,s, no faster than the measured latency, costs 3 points at the mild
tolerance and 8 at the strict while the gap to B1 (0.74 and 0.51) survives; a single plan per episode costs 13 and
18 (Table~\ref{tab:e4}). The result does not hinge on the unrealised 0.5\,s rate, but it does need replanning, and
drift of the state while a plan is computed remains unmodelled.

The committed strategy is not executed by prescribing the body. It enters the whole-body differential-IK QP as a
\emph{weighted twist reference}. At each 10\,ms step the QP solves for the 18 reduced velocities (pelvis twist,
waist and arm rates; the legs are eliminated by the fixed-feet constraint) and minimises three tasks: EE tracking
(weights 30 and 10 for position and orientation), strategy following (feed-forward plan velocity plus a gain of
5\,s$^{-1}$ on the offset to the plan, weight 5) and a posture task toward the plan (weight 1), plus a small motion
regulariser. Constraints are joint position and velocity limits, a pelvis workspace fence, pelvis speed caps of
0.15\,m/s and 0.6\,m/s$^2$ (below the 0.2\,m/s validity collapse of Sec.~\ref{sec:oracle}), and the CoM projection
staying inside the support polygon shrunk by 3\,cm. The EE tasks dominate, so the strategy decides how and when
the body is recruited while the tracker retains reactive authority over residual hand error. B1 is this same QP
with the home posture as reference and no strategy term; a \emph{hard} variant that prescribes the body exactly is
kept as an ablation.

\subsection{Execution layer and validation}\label{sec:exec}
All success numbers come from full-dynamics rollouts in MuJoCo~\cite{todorov2012mujoco} of the R1 standing free on
flat ground. The QP's joint and pelvis references are realised at 100\,Hz by a whole-body torque QP: floating-base
inverse dynamics with eight sole-corner contact forces in friction pyramids ($\mu=0.7$), the six base equations as
equalities, torque bounds, and acceleration-level PD on the joints and the planned pelvis pose. Joint-space PD
without this layer topples the robot in motion. Kinematics and dynamics quantities use Pinocchio~\cite{carpentier2019pinocchio}; QPs are
solved with OSQP~\cite{stellato2020osqp}. One rollout takes about 1.1\,s.

\section{Experiments}\label{sec:exp}
\textbf{Protocol.} \emph{All learning and evaluation use solvable episodes only}: an episode enters training or a
test set only if the oracle found at least one dynamically valid strategy for it, and 52\% of the generated suite
has none at the mild tolerance (Sec.~\ref{sec:e0}), so every success rate below is conditional on solvability.
AMBIT does not detect unsolvability online: when no sample passes verification the runtime falls back to the
reactive tracker (0.65 fallbacks per test episode), so an unsolvable task degrades to B1's behaviour. Models are
trained on 80\% of the 3000-episode main shard with three seeds; E0 is computed over the whole shard. E1 and E2 use
60 held-out episodes; E3 and E4 use 160 held-out episodes (480 episode-runs per method), and E4 is presented first
because E3 reuses its runtime. Runtime hyperparameters of the soft executor (verification tolerance and reference weight) were chosen
on 24 further held-out episodes disjoint from the test set, under a rule declared before the test run; the test
set was then run once. Every headline number is reported at the strict and the mild tolerance with 95\% bootstrap
confidence intervals over episodes. Baselines: B1, the reactive whole-body QP tracker; B2, the deterministic
regressor; B3, a per-pose generative whole-body IK sampler with smoothing (a CVAE over pelvis and waist states
conditioned on the current target, replanned at every 50\,ms step, following \cite{wbikdiffusion2026}); B4, a
test-time re-optimisation baseline that re-solves 16 whole-body QP trajectories from random-restart references with
full preview and filters them with the same constraint checkers (a finite optimiser, not an oracle).

\subsection{E0: the valid strategy set (H1)}\label{sec:e0}
Is the valid set genuinely multimodal? At the mild tolerance the oracle finds 11.3 valid candidates (of 41) per
episode, averaged over all episodes; 52\% of episodes have none. Among the solvable episodes there are 9.2 physically distinct strategies after
deduplication, and 92\% of them have at least two. The descriptor space has weak cluster structure (silhouette
0.33 at $k=2$) and an intrinsic dimension of about 4.3~\cite{facco2017twonn}: $\Zvalid$ is a diverse continuous
manifold, not a small set of discrete modes. H1 holds.

\begin{figure}[t]
\centering
\includegraphics[width=\linewidth]{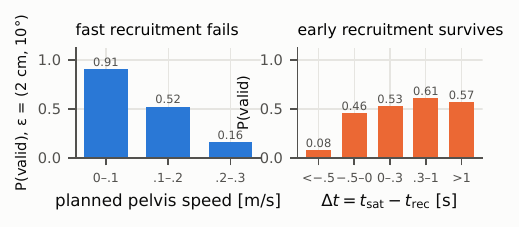}
\caption{E0. Probability that an oracle strategy is dynamically valid (strict tolerance) against its planned pelvis
speed (left) and anticipation margin (right): validity collapses above about 0.2\,m/s, and recruiting before the
arms saturate is valid more often than after.}
\label{fig:e0}
\end{figure}

\subsection{E1 and E2: mode averaging, precision and coverage (H2)}
\begin{figure}[t]
\centering
\includegraphics[width=\linewidth]{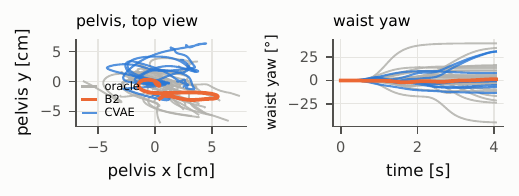}
\caption{E1, mode averaging on the held-out episode with the most valid strategies (lateral, asymmetric; 36 in
grey). The regressor's single output (B2, orange) runs down the middle of the fan and
leaves the waist near zero while valid strategies twist it from $-45^\circ$ to $+40^\circ$; CVAE samples (blue)
spread across the fan, though only half are valid at the mild tolerance (Table~\ref{tab:e2}). The flow is omitted: its heavy tails triple both axes.}
\label{fig:e1}
\end{figure}

Does regression onto that manifold average across it? Table~\ref{tab:e2} compares models trained on identical
data. Per-sample precision (the fraction of samples that are dynamically valid) is 0.35 for the regressor against
0.52 for the CVAE at the mild tolerance, with non-overlapping intervals; with 16 samples the CVAE has at least one
valid strategy in 82\% of episodes against 35\% for the single regressor output. The regressor covers 20\% of the
oracle's descriptor-space clusters, the CVAE 46\% and the flow 43\%; the flow has the lowest MMD but heavy tails
that hurt its precision. Qualitatively the regressor output lies between oracle modes (Fig.~\ref{fig:e1}). H2
holds. Evaluated on the unseen narrow-range shard, the CVAE's precision (0.60) and hit rate (0.85) do not degrade.

\begin{table}[t]
\centering\scriptsize
\caption{E1/E2: per-sample precision at the strict/mild/loose tolerances, probability that at least one of 16
samples is valid, and oracle-relative coverage at the mild tolerance; 60 held-out episodes, 3 seeds, 95\% CIs.}
\label{tab:e2}
\resizebox{\columnwidth}{!}{%
\begin{tabular}{lcccc}
\toprule
model & precision (2\,cm / 3\,cm / 5\,cm) & P($\geq$1 valid of 16) & coverage & MMD$^2$ \\
\midrule
B2 regressor & 0.15 / 0.35 [0.28, 0.42] / 0.54 & 0.35 [0.28, 0.42] & 0.20 & -- \\
CVAE (ours) & 0.33 / \textbf{0.52} [0.46, 0.57] / 0.72 & \textbf{0.82} [0.76, 0.87] & \textbf{0.46} & 0.11 \\
cond.\ flow & 0.20 / 0.34 [0.30, 0.39] / 0.52 & 0.69 [0.62, 0.76] & 0.43 & \textbf{0.08} \\
\bottomrule
\end{tabular}}
\end{table}

\subsection{E4: closed-loop success and anticipation (H3)}\label{sec:e4}
Does previewing the task let the planner succeed more often, and recruit earlier, than a reactive tracker? Table~\ref{tab:e4} reports the receding-horizon runtime on the 160 test episodes. Our planner with soft
execution and reactive fallback succeeds in 85\% of episode-runs at the mild tolerance and 74\% at the strict
tolerance, against 74\% and 51\% for the reactive tracker B1, with non-overlapping confidence intervals at both
levels. Paired per episode-run, the planner succeeds where B1 fails 69 times and the reverse happens 20 times
(127 vs.\ 18 at the strict tolerance). It is never worse than B1 in any task family and gains most where lateral
or low reaches force body recruitment (lateral: 0.80 vs.\ 0.59; low reach: 0.74 vs.\ 0.61; traversal: 0.95 vs.\
0.78).

\begin{figure*}[t]
\centering
\includegraphics[width=\textwidth]{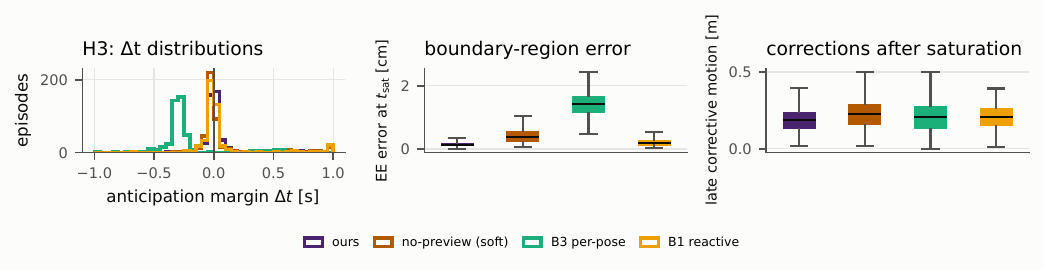}
\caption{E4. Left: anticipation margins $\Delta t=\tsat-\trec$ (positive: body moves before the arms saturate;
the outer bins collect $|\Delta t|>1$\,s). Middle: EE error around $\tsat$ (boxes: median and IQR). Right:
corrective body motion after saturation.}
\label{fig:e4}
\end{figure*}

\textbf{Anticipation.} The planner recruits the body before arm saturation in 48\% of episodes, against 35\% for
B1 and 30\% for the no-preview ablation on the identical runtime; per episode it recruits earlier than B1 in 34\%
of cases, at the same time in 58\% and later in 7\%. The margins are small: the median $\Delta t$ is close to
zero for every preview-conditioned method and for B1 (B3: $-0.29$\,s), and the planner's 90th percentile is
$+0.32$\,s. The cause is the physics of Sec.~\ref{sec:oracle}: saturation comes early (median $\tsat$ of
0.71\,s), valid recruitment must stay below 0.2\,m/s, so early recruitment has little room. Preview conditioning
is nevertheless what makes the planner work: the same model without preview succeeds in 56\% of episodes (soft
execution) or 13\% (hard execution). Binned by how fast the frozen-body error grows after saturation, the success
gain over B1 is uniform across bins (+6 to +11 points), while the anticipation gap is largest where the task
leaves time (73\% vs.\ 61\% of episodes in the lowest-demand tercile, 25\% vs.\ 17\% in the highest): the learned
reference recruits early when it can (Fig.~\ref{fig:e4}). H3 holds in success and in the share of early
recruitments, by small margins in time.

\textbf{Execution mode and baselines.} Hard execution, which prescribes the body exactly, reaches 81\% at the mild
tolerance but collapses to 29\% at the strict one, and anticipates marginally more often (50\%) at that price:
the arms alone absorb every residual and the median peak hand error is 2.4\,cm against 1.4\,cm for soft execution
and 1.2\,cm for B1, whose failures are late misses rather than peak error. The per-pose sampler B3 succeeds in
25\% of episodes and switches strategies 46 times per episode; blind to the future, it never anticipates (8\%).
Success is flat in the sample budget: $K=8,16,32,64$ give 0.80, 0.85, 0.80 and 0.78, so the remaining gap lies in
verification and execution, not in sampling.

\textbf{Is B1 a weak baseline?} B1 is the posture-regularised task-priority QP of AMO-style
pipelines~\cite{li2025amo} and shares every task weight, constraint, speed cap and the torque-controlled execution
layer with our executor; only the posture reference differs, so the gain is by construction a better reference.
Whether a non-learned reference could match it is open, but two results bound it: raising B1's pelvis speed cap
lowers its success (0.74 to 0.62, Sec.~\ref{sec:g1}) because faster recruitment is dynamically invalid, and B4, a
preview-based test-time optimiser of 16 whole-body QP trajectories, only matches our planner at 17$\times$ the
latency (Sec.~\ref{sec:e3}). A gradient-based whole-body model-predictive controller (MPC) with full-horizon preview was not run and is
the missing comparison.

\begin{table}[t]
\centering\scriptsize
\caption{E4, 160 held-out episodes $\times$ 3 seeds: success at the mild (3\,cm, 15$^\circ$; 95\% CI) and
strict (2\,cm, 10$^\circ$) tolerances, episodes recruiting before arm saturation, mean EE error around $\tsat$,
and strategy switches per episode.}
\label{tab:e4}
\resizebox{\columnwidth}{!}{%
\begin{tabular}{lccccc}
\toprule
method & mild [CI] & strict & anticip. & bnd.\ err & sw. \\
\midrule
ours (soft) + fallback & \textbf{0.85} [0.81, 0.88] & \textbf{0.74} & 48\% & 0.18\,cm & 1.7 \\
\quad replan every 1.0\,s & 0.85 [0.81, 0.88] & 0.73 & 48\% & 0.18\,cm & 1.1 \\
\quad replan every 1.5\,s (= latency) & 0.81 [0.78, 0.85] & 0.66 & 49\% & 0.18\,cm & 0.8 \\
\quad single plan at $t=0$ & 0.72 [0.68, 0.76] & 0.56 & 49\% & 0.18\,cm & 0.0 \\
ours (soft) & 0.82 [0.79, 0.86] & 0.66 & 48\% & 0.19\,cm & 2.6 \\
ours (hard) + fallback & 0.81 [0.77, 0.84] & 0.29 & 50\% & 0.36\,cm & 2.3 \\
no-preview (soft) & 0.56 [0.52, 0.61] & 0.45 & 30\% & 0.44\,cm & 2.5 \\
no-preview (hard) & 0.13 [0.10, 0.16] & 0.02 & 23\% & 1.62\,cm & 6.9 \\
B3 per-pose IK & 0.25 [0.22, 0.29] & 0.01 & 8\% & 1.50\,cm & 46.3 \\
B1 reactive QP & 0.74 [0.71, 0.78] & 0.51 & 35\% & 0.25\,cm & -- \\
\bottomrule
\end{tabular}}
\end{table}

\subsection{E3: zero-shot constraint shift (H4)}\label{sec:e3}
Can a distribution that stayed diverse honour constraints it never saw? At test time we impose five: a torso
obstacle that caps the pelvis height, a pelvis keep-out region in front, a reduced waist range, a forbidden
forward lean, and a 0.75\,kg payload on each wrist. The planner handles them by selection alone: strategy-space
filters, and for the payload a quasi-static CoM check inside verification. B1 is re-run with the shift as QP
bounds where the shift is expressible as bounds (keep-out, lean, waist range); under the torso obstacle and the
payload it runs unchanged and is only checked. B4 re-optimises 16 trajectories per episode under the
same checkers. Table~\ref{tab:e3} shows that the planner beats B1 on all five shifts (+6 to +18 points; +12 to +24 at the
strict tolerance), with disjoint 95\% intervals on four of the five at the mild tolerance and on all five at the
strict. It is at least as good as B4 on all five, at a latency of 1.5\,s against 25\,s with the $K$ candidates
verified in parallel. Sweeping the payload from 0.5 to 1.0\,kg per wrist, success degrades for every method (ours 0.83, 0.61, 0.46; B1
0.72, 0.43, 0.33; B4 0.74, 0.58, 0.47 at the mild tolerance), but the planner stays 11 to 18 points above the
load-unaware tracker at every load and never falls more than a point below B4. Success is counted only if the executed trajectory satisfies the shift constraints; under the same accounting the
hard-execution planner (first 40 episodes) reaches 0.72, 0.72, 0.72, 0.61 and 0.53, so soft execution contributes 6
to 15 points. H4 holds.

\begin{table}[t]
\centering\scriptsize
\caption{E3, post-selection success at the mild tolerance (strict in parentheses) on 160 held-out episodes
$\times$ 3 seeds; the shift is known only at test time. Wall time per plan, candidates verified in parallel:
1.5, 1.6, 1.7 and 25\,s for the four columns.}
\label{tab:e3}
\resizebox{\columnwidth}{!}{%
\begin{tabular}{lcccc}
\toprule
shift & ours (CVAE) & ours (flow) & B1 warm-started & B4 re-opt. \\
\midrule
torso obstacle & \textbf{0.86} (0.74) & 0.81 (0.72) & 0.74 (0.51) & 0.78 (0.58) \\
pelvis keep-out & \textbf{0.87} (0.75) & 0.82 (0.72) & 0.74 (0.51) & 0.78 (0.58) \\
waist range $\downarrow$ & \textbf{0.87} (0.75) & 0.82 (0.72) & 0.74 (0.51) & 0.78 (0.58) \\
no forward lean & \textbf{0.67} (0.58) & 0.66 (0.59) & 0.61 (0.46) & 0.67 (0.52) \\
payload 0.75\,kg & \textbf{0.61} (0.55) & 0.60 (0.55) & 0.43 (0.36) & 0.58 (0.46) \\
\bottomrule
\end{tabular}}
\end{table}


\subsection{Second platform: Unitree G1}\label{sec:g1}
\begin{figure*}[t]
\centering
\IfFileExists{figures/fig_platforms.pdf}{\includegraphics[width=0.68\textwidth]{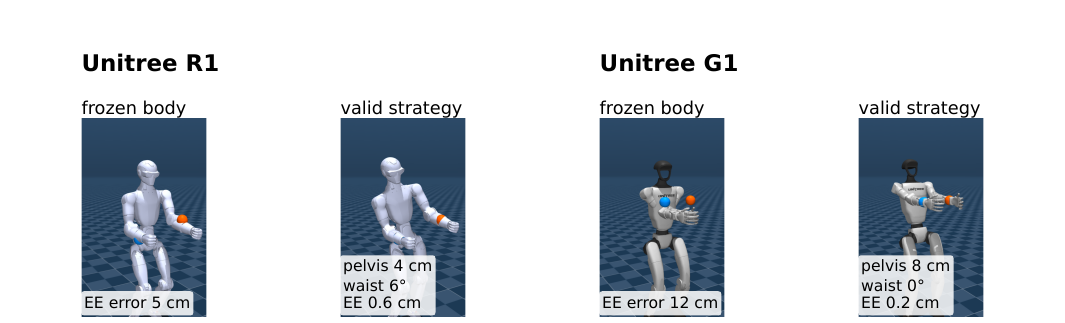}}{\todo{run the platforms figure script}}
\caption{One task per platform at the instant of the largest frozen-body miss (MuJoCo; a lateral task on the R1,
a traversal on the G1). Frozen body: the arms saturate and the hands miss their targets (spheres); one valid oracle
strategy brings both hands back within millimetres by moving the pelvis and, on the R1, the waist.}
\label{fig:platforms}
\end{figure*}

Are these findings about the R1 or about the problem? We re-ran the complete protocol on a second humanoid with
every hyperparameter transferred unchanged. The Unitree G1 has 6-DoF legs, a yaw/roll/pitch
waist and 7-DoF arms; fixing its waist pitch and two distal wrist joints places it in the kinematic class we study
(5-DoF arms, roll and yaw waist, 24 movable joints; Fig.~\ref{fig:platforms}) and differs otherwise: arm reach 0.41 against 0.36\,m, pelvis height 0.73 against 0.67\,m, mass 35.1 against 28.9\,kg, and
different joint ranges, torque limits and foot geometry. Only the robot description changes (3000 episodes, three seeds, 80 held-out
episodes for E4 and E3).

\textbf{The structural findings replicate.} The G1's valid set has the R1's shape: 9.5 valid strategies per
episode at the mild tolerance (R1: 11.3), 57\% of episodes with none (52\%), 9.0 distinct strategies per solvable
episode (9.2), 89\% of solvable episodes multimodal (92\%), intrinsic dimension 4.3 on both. Regression again
mode-averages: per-sample validity 0.28 [0.22, 0.35] for the regressor against 0.47 [0.42, 0.53] for the CVAE (R1:
0.35 against 0.52), whose 16 samples contain a valid strategy in 82\% of episodes on both robots.

\textbf{The closed-loop advantage is larger.} Our planner reaches 0.85 [0.80, 0.89] success against 0.53
[0.46, 0.59] for the reactive tracker at the mild tolerance and 0.77 against 0.45 at the strict one: a 32-point gap
where the R1 showed 11 (0.85 against 0.74). Paired per episode-run, the planner succeeds where the tracker fails 82
times and the reverse happens 4 times. The gap is largest where the body must move (traversals 0.95 against 0.36,
lateral reaches 0.72 against 0.33) and smallest where the arms nearly suffice (forward reaches 0.94 against 0.86).
Preview remains necessary: the same model without it reaches 0.47.

\textbf{Why the tracker degrades on the larger robot.} The G1 is 21\% heavier with 14\% longer arms, so targets sit
further from the body while the admissible pelvis speed (0.15\,m/s) is unchanged: recruiting after the arms saturate
is slower relative to the task. Accordingly 68\% of the tracker's failures are tolerance misses (90th-percentile
peak error 7.4\,cm), not falls, and across terciles of frozen-body error growth it falls from 0.67 to 0.41 while our
planner stays at 0.86, 0.85 and 0.84. Raising its speed cap to 0.25 or 0.40\,m/s does not help: success falls to
0.46 and the fall rate doubles to 36\%; faster recruitment is dynamically invalid (Fig.~\ref{fig:e0}).

\textbf{The planner's margin is largest under constraint shift.} Under the same five shifts our planner reaches
0.90, 0.90, 0.90, 0.78 and 0.77 (shifts in the order of Table~\ref{tab:e3}) against
0.53, 0.53, 0.53, 0.50 and 0.39 for the warm-started tracker and 0.65, 0.65, 0.63, 0.58 and 0.54 for the 24\,s
re-optimisation baseline B4, every interval disjoint from both: the planner beats B4 by 20 to 25 points, where
on the R1 it merely matched it, at 1.5\,s against 24\,s. We did not isolate why B4's random restarts fare
worse on the G1. Anticipation margins are larger on the G1 (69\% of episodes anticipatory against 48\%) but so is
the tracker's share (61\%), so that contrast stays modest.

\subsection{Hardware}
\label{sec:hardware}
We executed five selected strategies on the physical R1.
Command files are joint trajectories from the validated set, hash-sealed and re-checked at run time against inflated
margins (balance $\times 1.5$, joint limits $-10\%$, velocity $\times 0.5$, torque $\times 0.6$); four candidates
failed that gate and were not run. The hardware execution layer is joint-space PD, which cannot stabilise this
robot's unactuated base once the body moves (Sec.~\ref{sec:exec}), so the robot was supported: ankle-pitch torque of
at most $2.7$\,N\,m and tilt of at most $0.078$\,rad confirm the support carried the weight.

\begin{figure}[t]
\centering
\IfFileExists{figures/e5/fig_hardware.pdf}{\includegraphics[width=\linewidth]{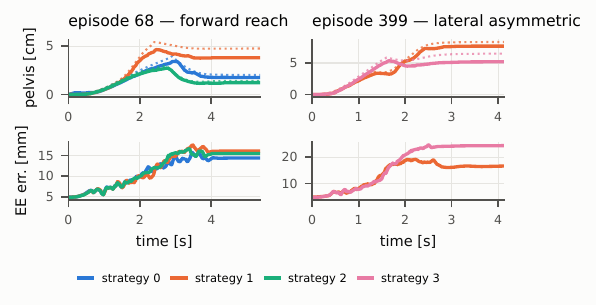}}{\todo{run experiments/e5\_hardware/make\_figures.py}}
\caption{Five strategies of two commands on the supported R1 (left: episode 68, three; right: episode 399, two).
Top: pelvis excursion from the encoders, feet as the fixed frame, measured (solid) against commanded
(dotted; commanded peaks in Table~\ref{tab:e5}). Bottom: end-effector error against the plan, worse hand.}
\label{fig:hardware}
\end{figure}

\textbf{Result.} Five strategies over two commands executed without intervention; Table~\ref{tab:e5} and
Fig.~\ref{fig:hardware} report the trajectory phase of each. Distinct strategies for the same command remain
distinct on hardware: the three strategies of episode~68 are commanded to peak pelvis excursions of $4.0$, $5.4$
and $3.0$\,cm and measure $3.5$, $4.7$ and $2.7$\,cm, an $8$--$17\%$ undershoot. End-effector error, from forward
kinematics of the encoders (no external tracker was available), has median $10.6$\,mm and reaches $24.7$\,mm against
$2$--$6$\,mm in simulation for the same files; median joint error is $6.8$\,mrad.

The gap is uncompensated gravity. The error does not track joint speed (correlation $-0.24$ to $+0.05$) but
tracks the arm's gravity load (correlation $0.96$ to $0.99$ with total arm torque) and persists after motion
stops. A position-controlled joint with stiffness $k_p$ holding a gravity torque $\tau$ settles at
an offset $\tau / k_p$: across the five runs that prediction gives a median arm-joint offset of
$10.1$--$11.3$\,mrad against $10.3$--$12.1$\,mrad observed, agreeing to within $13\%$. The torque QP used in
simulation cancels this term through inverse dynamics. Adding the model's arm and waist gravity torque as feed-forward on
the same position controller, still supported, halved the arm-joint error at the same pose (median $3.8$ to
$2.1$\,mrad, worst $17.5$ to $8.1$\,mrad). The hardware error belongs to the execution layer, not to the
strategies or the planner.

\textbf{What this establishes.} The runs show \emph{kinematic} execution: the trajectories lie inside the real
machine's envelope and the encoders reproduce the planned hand motion. Dynamic validity remains a simulation
result, and we saw the limitation directly: in one run the support was not carrying the weight and the robot
toppled as soon as the pelvis moved, as simulation predicts for joint PD. Running the torque QP on the robot agreed with the model (pelvis height to a millimetre, contact force within 1\,N
of the weight), but our Python loop could not hold its 10\,ms period (up to 38\,ms); free-standing execution is
future work. The accompanying video is provided as an ancillary file of this preprint.

\begin{table}[t]
\centering\scriptsize
\caption{Hardware execution (supported robot), trajectory phase. Pelvis: commanded excursion. End-effector error:
forward kinematics of the encoders against the plan, both hands pooled.}
\label{tab:e5}
\IfFileExists{tables/e5_hardware.tex}{\resizebox{\columnwidth}{!}{\begin{tabular}{ll r r r r r r r r}
\toprule
ep. & str. & pelvis & dur. & \multicolumn{2}{c}{joint err.\ [mrad]} & \multicolumn{3}{c}{EE pos.\ err.\ [mm]} & rot.\ [deg] \\
\cmidrule(lr){5-6}\cmidrule(lr){7-9}
 & & [cm] & [s] & med. & max & mean & p95 & max & mean \\
\midrule
68 & 0 & 4.0 & 5.4 & 5.6 & 40 & 10.5 & 14.4 & 15.8 & 2.35 \\
68 & 1 & 5.4 & 5.4 & 7.3 & 43 & 10.3 & 16.3 & 17.6 & 2.44 \\
68 & 2 & 3.0 & 5.4 & 7.2 & 46 & 10.3 & 15.6 & 16.7 & 2.30 \\
399 & 1 & 8.3 & 4.1 & 6.6 & 65 & 13.1 & 18.4 & 19.2 & 2.65 \\
399 & 3 & 6.5 & 4.1 & 7.9 & 58 & 13.6 & 24.3 & 24.7 & 2.84 \\
\bottomrule
\end{tabular}
}}{\todo{run experiments/e5\_hardware/make\_figures.py}}
\end{table}

\section{Limitations and Conclusion}
All results are conditional on solvability (Sec.~\ref{sec:exp}). Anticipation margins are tens of milliseconds.
Planning latency (1.5\,s) exceeds the 0.5\,s replan interval and is not simulated (Sec.~\ref{sec:runtime}). No
preview-based whole-body MPC baseline was run; B4 is the nearest comparison. We did not
run the regressor under the constraint shifts, so H4's mechanism is inferred from E1/E2 rather than isolated.
Strict-tolerance numbers are tabulated without confidence intervals. The hardware runs use supported position
control and establish kinematic realisability, not balance. Torque limits are taken from the URDF and MJCF
models. Stepping is out of scope.

Body recruitment for bimanual tracking on a humanoid is a multimodal, temporally extended decision. The set of
valid recruitment strategies is a diverse continuous manifold, and regressing onto it mode-averages. AMBIT, a
generative planner whose samples serve as references for a standard whole-body tracker, is significantly more
successful than the tracker alone, recruits the body before saturation in more episodes (by small margins in
time), and honours constraints it never saw at training time by selection alone; it matches a 17$\times$ more
expensive test-time re-optimisation baseline on one robot and beats it on a second, heavier one.
The executor design matters as much as the distribution: the tracker keeps authority over residual hand error
while the strategy decides when and how the body moves.

\bibliographystyle{IEEEtran}
\bibliography{refs}

@IEEEtranBSTCTL{IEEEexample:BSTcontrol,
  CTLuse_forced_etal       = "yes",
  CTLmax_names_forced_etal = "6",
  CTLnames_show_etal       = "1"
}

@article{ames2022ikflow,
  title   = {{IKFlow}: Generating Diverse Inverse Kinematics Solutions},
  author  = {Ames, Barrett and Morgan, Jeremy and Konidaris, George},
  journal = {IEEE Robotics and Automation Lett.},
  volume  = {7}, number = {3}, pages = {7177--7184}, year = {2022}
}

@article{limoyo2023ggik,
  title   = {Generative Graphical Inverse Kinematics},
  author  = {Limoyo, Oliver and Mari{\'c}, Filip and Giamou, Matthew and Alexson, Petra and Petrovi{\'c}, Ivan and Kelly, Jonathan},
  journal = {IEEE Trans. Robotics},
  volume  = {41}, pages = {1002--1018}, year = {2025}
}

@misc{ikdiffuser2025,
  title  = {{IKDiffuser}: A Diffusion-based Generative Inverse Kinematics Solver for Kinematic Trees},
  author = {Zhang, Zeyu and Jiao, Ziyuan},
  year   = {2025},
  note   = {arXiv:2506.13087}
}

@misc{wbikdiffusion2026,
  title  = {Whole-Body Inverse Kinematics with Graph Diffusion},
  author = {Huang, Helong and Tan, Kai and Wen, Feng and Huang, Guowei and Quan, Xingyue},
  year   = {2026},
  note   = {arXiv:2606.00086}
}

@misc{li2025amo,
  title  = {{AMO}: Adaptive Motion Optimization for Hyper-Dexterous Humanoid Whole-Body Control},
  author = {Li, Jialong and Cheng, Xuxin and Huang, Tianshu and Yang, Shiqi and Qiu, Ri-Zhao and Wang, Xiaolong},
  year   = {2025},
  note   = {arXiv:2505.03738}
}

@inproceedings{chi2023diffusionpolicy,
  title     = {Diffusion Policy: Visuomotor Policy Learning via Action Diffusion},
  author    = {Chi, Cheng and Feng, Siyuan and Du, Yilun and Xu, Zhenjia and Cousineau, Eric and Burchfiel, Benjamin and Song, Shuran},
  booktitle = {Proc. Robotics: Science and Systems (RSS)},
  year      = {2023}
}

@inproceedings{sohn2015cvae,
  title     = {Learning Structured Output Representation using Deep Conditional Generative Models},
  author    = {Sohn, Kihyuk and Lee, Honglak and Yan, Xinchen},
  booktitle = {Proc. NeurIPS},
  year      = {2015}
}

@inproceedings{dinh2017realnvp,
  title     = {Density Estimation using {Real NVP}},
  author    = {Dinh, Laurent and Sohl-Dickstein, Jascha and Bengio, Samy},
  booktitle = {Proc. ICLR},
  year      = {2017}
}

@article{gretton2012mmd,
  title   = {A Kernel Two-Sample Test},
  author  = {Gretton, Arthur and Borgwardt, Karsten M. and Rasch, Malte J. and Sch{\"o}lkopf, Bernhard and Smola, Alexander},
  journal = {J. Machine Learning Research},
  volume  = {13}, pages = {723--773}, year = {2012}
}

@article{facco2017twonn,
  title   = {Estimating the intrinsic dimension of datasets by a minimal neighborhood information},
  author  = {Facco, Elena and d'Errico, Maria and Rodriguez, Alex and Laio, Alessandro},
  journal = {Scientific Reports},
  volume  = {7}, number = {1}, pages = {12140}, year = {2017}
}

@inproceedings{todorov2012mujoco,
  title     = {{MuJoCo}: A physics engine for model-based control},
  author    = {Todorov, Emanuel and Erez, Tom and Tassa, Yuval},
  booktitle = {Proc. IEEE/RSJ Int. Conf. Intelligent Robots and Systems (IROS)},
  pages     = {5026--5033}, year = {2012}
}

@inproceedings{carpentier2019pinocchio,
  title     = {The {Pinocchio} {C++} library: A fast and flexible implementation of rigid body dynamics algorithms and their analytical derivatives},
  author    = {Carpentier, Justin and Saurel, Guilhem and Buondonno, Gabriele and Mirabel, Joseph and Lamiraux, Florent and Stasse, Olivier and Mansard, Nicolas},
  booktitle = {Proc. IEEE/SICE Int. Symp. System Integration (SII)},
  year      = {2019}
}

@article{stellato2020osqp,
  title   = {{OSQP}: an operator splitting solver for quadratic programs},
  author  = {Stellato, Bartolomeo and Banjac, Goran and Goulart, Paul and Bemporad, Alberto and Boyd, Stephen},
  journal = {Math. Programming Computation},
  volume  = {12}, number = {4}, pages = {637--672}, year = {2020}
}

@article{kanoun2011prioritized,
  title   = {Kinematic Control of Redundant Manipulators: Generalizing the Task-Priority Framework to Inequality Task},
  author  = {Kanoun, Oussama and Lamiraux, Florent and Wieber, Pierre-Brice},
  journal = {IEEE Trans. Robotics},
  volume  = {27}, number = {4}, pages = {785--792}, year = {2011}
}

@article{sentis2005synthesis,
  title   = {Synthesis of Whole-Body Behaviors through Hierarchical Control of Behavioral Primitives},
  author  = {Sentis, Luis and Khatib, Oussama},
  journal = {Int. J. Humanoid Robotics},
  volume  = {2}, number = {4}, pages = {505--518}, year = {2005}
}

@misc{cheng2024exbody,
  title  = {Expressive Whole-Body Control for Humanoid Robots},
  author = {Cheng, Xuxin and Ji, Yandong and Chen, Junming and Yang, Ruihan and Yang, Ge and Wang, Xiaolong},
  year   = {2024},
  note   = {arXiv:2402.16796}
}

@misc{fu2024humanplus,
  title  = {{HumanPlus}: Humanoid Shadowing and Imitation from Humans},
  author = {Fu, Zipeng and Zhao, Qingqing and Wu, Qi and Wetzstein, Gordon and Finn, Chelsea},
  year   = {2024},
  note   = {arXiv:2406.10454}
}

@misc{he2024omnih2o,
  title  = {{OmniH2O}: Universal and Dexterous Human-to-Humanoid Whole-Body Teleoperation and Learning},
  author = {He, Tairan and Luo, Zhengyi and He, Xialin and Xiao, Wenli and Zhang, Chong and Zhang, Weinan and Kitani, Kris and Liu, Changliu and Shi, Guanya},
  year   = {2024},
  note   = {arXiv:2406.08858}
}

@article{yamane2003dynamicsfilter,
  title   = {Dynamics Filter---Concept and Implementation of Online Motion Generator for Human Figures},
  author  = {Yamane, Katsu and Nakamura, Yoshihiko},
  journal = {IEEE Trans. Robotics and Automation},
  volume  = {19}, number = {3}, pages = {421--432}, year = {2003}
}

@inproceedings{yamane2009tracking,
  title     = {Simultaneous Tracking and Balancing of Humanoid Robots for Imitating Human Motion Capture Data},
  author    = {Yamane, Katsu and Hodgins, Jessica K.},
  booktitle = {Proc. IEEE/RSJ Int. Conf. Intelligent Robots and Systems (IROS)},
  pages     = {2510--2517}, year = {2009}
}

@inproceedings{kajita2003rmc,
  title     = {Resolved Momentum Control: Humanoid Motion Planning Based on the Linear and Angular Momentum},
  author    = {Kajita, Shuuji and Kanehiro, Fumio and Kaneko, Kenji and Fujiwara, Kiyoshi and Harada, Kensuke and Yokoi, Kazuhito and Hirukawa, Hirohisa},
  booktitle = {Proc. IEEE/RSJ Int. Conf. Intelligent Robots and Systems (IROS)},
  pages     = {1644--1650}, year = {2003}
}

@inproceedings{nakaoka2003dances,
  title     = {Generating Whole Body Motions for a Biped Humanoid Robot from Captured Human Dances},
  author    = {Nakaoka, Shin'ichiro and Nakazawa, Atsushi and Yokoi, Kazuhito and Hirukawa, Hirohisa and Ikeuchi, Katsushi},
  booktitle = {Proc. IEEE Int. Conf. Robotics and Automation (ICRA)},
  pages     = {3905--3910}, year = {2003}
}

@article{dalibard2013wbmp,
  title   = {Dynamic Walking and Whole-Body Motion Planning for Humanoid Robots: An Integrated Approach},
  author  = {Dalibard, S{\'e}bastien and El Khoury, Antonio and Lamiraux, Florent and Nakhaei, Alireza and Ta{\"i}x, Michel and Laumond, Jean-Paul},
  journal = {Int. J. Robotics Research},
  volume  = {32}, number = {9--10}, pages = {1089--1103}, year = {2013}
}

@incollection{yoshida2017wbmp,
  title     = {Whole-Body Motion Planning},
  author    = {Yoshida, Eiichi and Kanehiro, Fumio and Laumond, Jean-Paul},
  booktitle = {Humanoid Robotics: A Reference},
  editor    = {Goswami, Ambarish and Vadakkepat, Prahlad},
  publisher = {Springer}, address = {Dordrecht},
  pages     = {1575--1599}, year = {2017}
}

\end{document}